\documentclass[
dvipsnames, table,   %
format=sigconf,
anonymous=false,     %
authorversion=true, %
]{acmart}

\usepackage{siunitx}
\newcolumntype{v}{>{\small}S}
\usepackage{tikz}
\usetikzlibrary{arrows, decorations.markings}

\usepackage{xspace}

\usepackage{algcompatible, algorithm}
\usepackage[noend]{algpseudocode}

\newcommand{\best}{{\cellcolor[gray]{0.75}}}
\newcommand{\statsimilar}{{\cellcolor[gray]{0.9}}}

\newcommand\mX{\mathcal{X}}

\newcommand\Real{\mathbb{R}}
\newcommand\Normal{\mathcal{N}}

\newcommand{\bx}{\mathbf{x}}
\newcommand{\bff}{\mathbf{f}}
\newcommand{\bmm}{\mathbf{m}}
\newcommand{\bww}{\mathbf{w}}
\newcommand{\bhh}{\mathbf{h}}

\newcommand{\bI}{\mathbf{I}}
\newcommand{\bz}{\mathbf{z}}
\newcommand{\bone}{\mathbf{1}}

\newcommand{\xnext}{\bx'}
\newcommand{\bkappa}{\boldsymbol{\kappa}}
\newcommand{\btheta}{\boldsymbol{\theta}}

\newcommand{\fstar}{f^\star}

\newcommand{\given}{\,|\,}
\DeclareMathOperator{\LatinHypercubeSampling}{SpaceFillingSampling}

\DeclareMathOperator*{\argmax}{\arg\!\max}
\DeclareMathOperator*{\argmin}{\arg\!\min}
\DeclareMathOperator*{\median}{median}

\newcommand{\mGP}{\ensuremath{\mathcal{GP}}\xspace}

\newcommand{\Data}{\mathcal{D}}
\newcommand{\Matern}{Mat{\'e}rn\xspace}

\newcommand*{\eg}{e.g.\@\xspace}
\newcommand*{\ie}{i.e.\@\xspace}

\newcommand{\pushfour}{\ensuremath{\textsc{push4}}\xspace}
\newcommand{\pusheight}{\ensuremath{\textsc{push8}}\xspace}

\copyrightyear{2020} 
\acmYear{2020} 
\setcopyright{acmlicensed}
\acmConference[GECCO '20 Companion]{Genetic and Evolutionary Computation Conference Companion}{July 8--12, 2020}{Cancún, Mexico}
\acmBooktitle{Genetic and Evolutionary Computation Conference Companion (GECCO '20 Companion), July 8--12, 2020, Cancún, Mexico}
\acmPrice{15.00}
\acmDOI{10.1145/3377929.3398118}
\acmISBN{978-1-4503-7127-8/20/07}

\begin{document}
\title[What do you Mean? The Role of the Mean Function in Bayesian Optimisation]
{What do you Mean? \\The Role of the Mean Function in Bayesian Optimisation}

\author{George {De Ath}}
\email{g.de.ath@exeter.ac.uk}
\orcid{0000-0003-4909-0257}
\affiliation{%
  \department{Department of Computer Science}
  \institution{University of Exeter}
  \city{Exeter}
  \country{United Kingdom}
}

\author{Jonathan E. Fieldsend}
\email{j.e.fieldsend@exeter.ac.uk}
\orcid{0000-0002-0683-2583}
\affiliation{%
  \department{Department of Computer Science}
  \institution{University of Exeter}
  \city{Exeter}
  \country{United Kingdom}
}

\author{Richard M. Everson}
\email{r.m.everson@exeter.ac.uk}
\orcid{0000-0002-3964-1150}
\affiliation{%
  \department{Department of Computer Science}
  \institution{University of Exeter}
  \city{Exeter}
  \country{United Kingdom}
}



\begin{abstract}
Bayesian optimisation is a popular approach for optimising expensive 
black-box functions. The next location to be evaluated is selected via
maximising an acquisition function that balances exploitation and exploration.
Gaussian processes, the surrogate models of choice in Bayesian optimisation, 
are often used with a constant prior mean function equal to the arithmetic mean
of the observed function values. We show that the rate of convergence can
depend sensitively on the choice of mean function.  We empirically investigate
8 mean functions (constant functions equal to the arithmetic mean,
minimum, median and maximum of the observed function evaluations, linear, 
quadratic polynomials, random forests and RBF networks), using 10 synthetic 
test problems and two real-world problems, and using the Expected Improvement
and Upper Confidence Bound acquisition functions.

We find that for design dimensions $\ge5$ using a constant mean function equal 
to the worst observed quality value is consistently the best choice on the 
synthetic problems considered. We argue that this worst-observed-quality 
function promotes exploitation leading to more rapid convergence. However, for
the real-world tasks the more complex mean functions capable of modelling the
fitness landscape may be effective, although there is no clearly optimum
choice.
\end{abstract}

\maketitle

\section{Introduction}
\label{sec:intro}
Bayesian optimisation (BO) is a popular approach for optimising expensive
(in terms of time and/or money) black-box functions that have no closed-form
expression or derivative information \citep{snoek:practical, shahriari:ego}. It
is a surrogate-based modelling approach that employs a probabilistic model
built with previous function evaluations. The Gaussian process (GP) model
typically used in BO provides a posterior predictive distribution that models
the target function in question and quantifies the amount of predictive
uncertainty. A GP is a collection of random variables,  any finite
number of which have a joint Gaussian distribution \citep{rasmussen:gpml}. It 
can be fully specified by its mean function and kernel function (also 
known as a covariance function) \citep{rasmussen:gpml}. The kernel and mean
functions may be regarded as specifying a Bayesian prior on the functions
from which the data are generated. The kernel function 
describes the structure, such as the smoothness and amplitude, of the functions
that can be modelled, while the mean function specifies the prior expected
value of the function at any location
\citep{shahriari:ego}.

In BO, the location that maximises an acquisition function (or infill 
criterion) is chosen as the next location to be expensively evaluated. 
Acquisition functions combine the surrogate model's predictions and the 
uncertainty about its prediction to strike a balance between myopically 
exploiting areas of design space that are predicted to yield good-quality
solutions and exploring regions that have high predicted uncertainty.

In the literature, many acquisition functions have been proposed in a number of works 
\citep{mockus:ei, srinivas:ucb, hernandez:PES, wang:MES, death:egreedy, 
kushner:ego}, and, in general, no one strategy has been shown to be 
all-conquering due to the no free lunch theorem \citep{wolpert:nofreelunch}.
However, recent works have shown that purely exploiting the surrogate model 
becomes a more effective strategy as the dimensionality of the problem 
increases \citep{death:egreedy, rehbach:ei_pv}. Similarly, the role of the 
kernel function in BO has been investigated \citep{chugh:covariance, 
lindauer:bopt_assess, mukhopadhyay:kriging, acar:cov, palar:kernels}. One of
the most popular kernels is the radial basis function (also known as the
squared exponential kernel) \citep{snoek:practical}. However, it is generally
regarded as being too smooth for real-world functions
\citep{stein:interpolation, rasmussen:gpml}, and the \Matern family of kernels
is often preferred.

Contrastingly, little attention has been paid to the role of the mean function
in BO, with general practise being to use a constant value of zero
\citep{wang:MES, hernandez:PES, death:egreedy}, although the constant value can
also be inferred from the data \citep{shahriari:ego}. In general regression
tasks, other mean functions have been considered, such as polynomials
\citep{blight:polymean, kennedy:bayescalib, williamson:hmbias}, and, more
recently, non-parametric methods such as neural networks 
\citep{iwata:meanNN, fortuin:meandeepnn}. In light of the lack of previous work into
the role of the mean function in BO, we investigate the effect of different 
mean functions in BO in terms of both the convergence rate of the optimisation
and quality of the best found solution. Specifically, we compare the 
performance of using different constant values, linear and quadratic functions,
as well as using random forests and radial basis function networks.

Our main contributions can be summarised as follows:
\begin{itemize}
\item We provide the first empirical study of the effect of using different Gaussian 
      process mean functions in Bayesian optimisation.
\item We evaluate  eight mean functions on ten well-known, synthetic test 
      problems and two real-world applications. This assessment is on a range of design dimensions (2 to 10) and for two popular 
acquisition functions.
    \item We show empirically, and explain in terms of the exploration
      versus exploitation trade-off, that choosing the mean function to be
      the constant function equal to the worst-seen so far evaluation of
      the objective function is consistently no worse and often superior
      to other choices of  mean function. 
\end{itemize}

We begin in Section~\ref{sec:bo} by briefly reviewing Bayesian optimisation.
In Section~\ref{sec:gp} we review Gaussian processes, paying particular
attention to the mean function and introduce the various mean functions we
evaluate in this work. Extensive empirical experimentation is carried out on
well-known test problems and a two real-world applications in
Section~\ref{sec:results}. We finish with concluding remarks in
Section~\ref{sec:conc}.

\section{Bayesian Optimisation}
\label{sec:bo}
\begin{algorithm} [t!]
  \caption{Sequential Bayesian optimisation.}
  \label{alg:bo}
 \begin{algorithmic}[]
 	\State \textbf{Inputs:}
 	\State {\setlength{\tabcolsep}{2pt}%
 	        \begin{tabular}{c p{2pt} l}
 			$M$ &:& Number of initial samples \\
 			$T$ &:& Budget on the number of expensive evaluations
 			\end{tabular}
 	       }%
  \end{algorithmic}
  \medskip
 
  \begin{algorithmic}[1]
    \Statex \textbf{Steps:}
	\State $X \gets \LatinHypercubeSampling(\mX, M)$ \label{alg:bo:lhs}
	        \Comment{\small{Initial samples}}
        \For{$t = 1 \rightarrow M$}
        \State $f_t \gets f(\bx_t)$
        	\Comment{\small{Expensively evaluate all initial samples}}
        \EndFor
	\State $\Data \gets \{(\bx_t, f_t)\}_{t=1}^M$ \label{alg:bo:trainingdata}
	\For{$t = M+1 \rightarrow T$}
        \State  $\btheta \gets  \text{Train\mGP}(\Data)$ \label{alg:bo:train}
        	\Comment{\small{Train a GP model}}
        \State $\xnext \gets \argmax_{\bx \in \mX}~ \alpha (\bx, \btheta)$ \label{alg:bo:xnext}
        	\Comment{\small{Maximise infill criterion}}
        \State $f' \gets f(\xnext)$ \label{alg:bo:eval}
	        \Comment{\small{Expensively evaluate $\xnext$}}
		\State $\Data \gets \Data \cup \{(\xnext, f')\}$ \label{alg:bo:augment}
			\Comment{\small{Augment training data}}
	\EndFor
	\State \Return $\Data$
  \end{algorithmic}
\end{algorithm}
Bayesian optimisation (BO), also known as Efficient Global Optimisation (EGO),
is a surrogate-assisted global search strategy that sequentially samples
design space at likely locations of the global optimum, taking into account both 
the surrogate model's prediction $\mu(\bx)$ and the associated prediction 
uncertainty $\sigma(\bx)$  \citep{jones:ego}. See 
\citep{shahriari:ego, brochu:tutorial, frazier:tutorial} for comprehensive
reviews of BO. Without loss of generality, we can define the problem of 
finding a global minimum of an unknown objective function
$f : \Real^d \mapsto \Real$ as
\begin{equation}
\min_{\bx \in \mX} f(\bx),
\end{equation}
where $\mX \subset \Real^d$ is the feasible design space of interest.
We assume that $f$ is a black-box function, \ie it has no simple
closed form, but that we can have access the results of its evaluations
$f(\bx)$ at any location $\bx \in \mX$, although evaluating $f(\bx)$ is
expensive so that the number of evaluations required to locate the global
optimum should be minimised.

Algorithm~\ref{alg:bo} outlines the BO procedure. 
It starts (line~\ref{alg:bo:lhs}) with a space filling design, typically 
Latin hypercube sampling \citep{mckay:lhs}, of the feasible space. These
samples $X = \{\bx_t\}_{t=1}^M$ are then expensively evaluated with the
function $f_t = f(\bx_t)$, and a training dataset $\Data$ is constructed
(line~\ref{alg:bo:trainingdata}). Then, at each iteration of the sequential
algorithm, a regression model, usually a Gaussian process ($\mGP$), is 
constructed and trained (line~\ref{alg:bo:train}) using the current training 
data. The choice of where next to expensively evaluate is determined by
maximising an acquisition function (or infill criterion) $\alpha(\bx)$ which 
balances the exploitation of regions of design space that are predicted to
yield good-quality solutions and exploration of regions of space where the
predictive uncertainty is high. The design $\xnext$ maximising 
$\alpha(\bx)$ is expensively evaluated and the training data is subsequently 
augmented (lines \ref{alg:bo:xnext} to \ref{alg:bo:augment}). This process is 
then repeated until the budget has been expended.

Two of the most popular acquisition functions are Expected Improvement
(EI) \citep{mockus:ei} and Upper Confidence Bound (UCB) \citep{srinivas:ucb}.
EI measures the positive predicted improvement over the best solution 
 evaluated thus far, $\fstar$:
\begin{equation}
\label{eqn:ei}
\alpha_{EI}(\bx) = \sigma(\bx) \left( s \Phi(s) + \phi(s) \right),
\end{equation}
where $s = (\fstar - \mu(\bx)) / \sigma(\bx))$ is the predicted improvement at
$\bx$ normalised by the uncertainty, and $\Phi(\cdot)$ and $\phi(\cdot)$ are
the Gaussian cumulative density and probability density functions respectively.
UCB is a weighted sum of the mean prediction and its associated uncertainty:
\begin{equation}
\label{eqn:ucb}
\alpha_{UCB}(\bx) = -\left( \mu(\bx) - \sqrt{\beta_t} \sigma(\bx) \right),
\end{equation}
where $\beta_t \geq 0$ is a weight that  depends on the number of
function evaluations $t$ and  explicitly controls the 
exploitation vs. exploitation trade-off. Note that both EI and UCB are
presented here in the form used for minimisation. While  other 
acquisition functions have been proposed, such as probability
of improvement \citep{kushner:ego} and  entropy-based methods such as 
predictive entropy search \citep{hernandez:PES} and max-value entropy search 
\citep{wang:MES}, we limit our investigation  to the commonly used EI and UCB to allow the focus on the mean functions themselves.

\begin{figure*}[t]
    \includegraphics[width=\textwidth, clip, trim={0 0 0 0}]{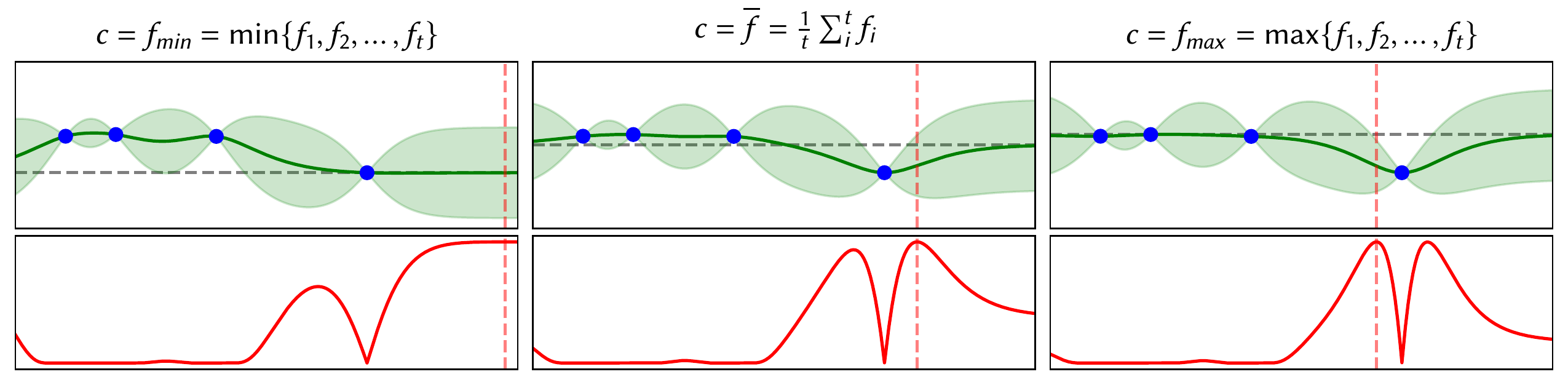}
    \caption{GP models with identical training data (blue) and kernel 
    hyperparameters, along with different constant mean functions. The mean 
    function used in each GP is shown (grey, dashed) and corresponds to the value of the
    smallest (\textit{left}), arithmetic mean  (\textit{centre}), and largest 
    (\textit{right}) seen function values. The lower row shows the corresponding EI
    and its maximiser (red dashed). Note how the $\min$ and $\max$ models 
    lead EI to prefer exploring and exploiting respectively.}
    \label{fig:mean_func_example}
    \end{figure*}
\section{Gaussian Processes}
\label{sec:gp}
Gaussian processes  are a common choice of surrogate model due to their 
strengths in uncertainty quantification and function approximation
\citep{shahriari:ego, rasmussen:gpml}. A GP is a collection of random 
variables, and any finite number of these have a joint Gaussian distribution 
\citep{rasmussen:gpml}. We can define a Gaussian process prior over $f$ to be
$\mGP(m(\bx), \kappa(\bx, \bx' \given \btheta))$, where 
$m(\cdot)$ is the mean function and $\kappa(\cdot, \cdot \given \btheta)$ is 
the kernel function (covariance function) with hyperparameters $\btheta$. 
Given data consisting of $f$ evaluated at $t$ sampled locations
$\Data = \{(\bx_n, f_n \triangleq f(\bx_n)) \}_{n=1}^t$, the posterior 
distribution of $f$ at a given location $\bx$ is Gaussian:
\begin{equation}
\label{eqn:gp:density}
p ( f \given \bx, \Data, \btheta) = \Normal ( f \given \mu(\bx), \sigma^2(\bx) )
\end{equation}
with posterior mean and variance
\begin{align}
\mu(\bx \given \Data, \btheta)
  &= m(\bx) + \bkappa( \bx, X ) K^{-1} ( \bff - \bmm )
  \label{eqn:gp:pred}\\
\sigma^2(\bx \given \Data, \btheta)
  &= \kappa(\bx, \bx) - \bkappa( \bx, X)^\top K^{-1} \kappa(X, \bx).
  \label{eqn:gp:var}
\end{align}
Here $X \in \Real^{t \times d}$ is the matrix of design locations,
$\bff = \{f_1, f_2, \dots, f_t\}$ is the corresponding vector of true function
evaluations and $\bmm = \{m(\bx_1), m(\bx_1), \dots, m(\bx_t)\}$ is the vector
comprised of the mean function at the  design locations. The kernel
matrix $K \in \Real^{t \times t}$ is given by 
$K_{ij} = \kappa(\bx_i, \bx_j \given \btheta)$ and $\bkappa(\bx, X)$ is 
given by $[\bkappa(\bx, X)]_i = \kappa(\bx, \bx_i \given \btheta)$. 
In this work we use an isotropic \Matern~$5 / 2$ kernel:

\begin{equation}
\label{eqn:mat52}
\kappa(\bx, \xnext \given \btheta) = \theta_0 \left( 1 + \sqrt{5}r + \frac{5}{3}r^2 \right)
\exp\left(-\sqrt{5}r\right),
\end{equation}
where $r = \theta_1 \lVert \bx - \xnext \rVert$ and 
$\btheta = [\theta_0, \theta_1]$, as recommended for modelling realistic
functions \citep{snoek:practical}. The kernel's hyperparameters $\btheta$ are 
learnt via maximising the log marginal likelihood (up to a constant):
\begin{equation}
\label{eqn:mll}
\log p(\bff \given X, \btheta) = - \frac{1}{2} \log \lvert K \rvert
- \frac{1}{2} \left( \bff - \bmm \right)^\top K^{-1} \left( \bff - \bmm \right)
\end{equation}
using a multi-restart strategy \citep{shahriari:ego} with L-BFGS-B
\citep{byrd:lbfgs}. Henceforth, we drop the explicit dependencies on the data 
$\Data$ and the kernel hyperparameters $\btheta$ for notational simplicity.

\subsection{Mean Functions}
\label{sec:meanfuncs}

It is well-known that the posterior prediction \eqref{eqn:gp:density} of the
GP reverts to the prior as the distance of $\bx$ from the observed data
increases, \ie as $\min_{\bx_i \in \Data} \lVert \bx - \bx_i \rVert \rightarrow
\infty$. In particular, the posterior mean reverts to the prior mean, \ie
$\mu(\bx) \approx m(\bx)$; and the posterior variance approaches the prior
variance, $\theta_0$ for the \Matern kernel. 
The upper row of Figure~\ref{fig:mean_func_example} illustrates this effect 
for three different constant mean values: the best, average (arithmetic
mean), and worst function
values observed thus far. Note how the predicted values of the GP tend to
 the mean function (dashed) as the distance from the nearest
evaluated location (blue) increases. The figure also shows another important, 
and often overlooked, aspect of the mean function: its effect on the 
acquisition function (lower). In this case three different locations maximise 
EI for the three mean functions, and it is not known \textit{a priori}
which location is preferable.

Practitioners of BO usually standardise the observations $f_t$ before fitting
the GP at each iteration, \ie they subtract the mean of the observations
and divide by the standard deviation of the relevant feature/variable after
which the mean function is taken as the constant function equal to zero;
that is, the effective mean function is the constant function equal to the
arithmetic mean of the observed function values.  In addition, the prior
variance of the GP is matched to the observed variance of the function
values.  Although the standardisation is not usually 
discussed in the literature, it is commonplace in standard BO libraries,
\eg GPyOpt~\citep{gpyopt}, BoTorch~\citep{botorch} and
Spearmint\footnote{\url{https://github.com/HIPS/Spearmint}}. 
It is, however, unclear whether this is the best choice for BO  or
whether a different mean function may be preferable. We now introduce
the mean functions that will be evaluated in this work.

Here, we consider the mean function to be a set of basis functions $\bhh(\bx)$
with corresponding weights $\bww$ \citep{rasmussen:gpml}:
\begin{equation}
\label{eqn:mf}
m(\bx) = \bhh(\bx)^\top \bww.
\end{equation}
A constant mean function with value $c$, for example, can be written
as $\bhh(\bx) = c\bone$ with $\bww = \bone$. In addition to the standard constant
function equal to the arithmetic mean of the data 
$c= \bar{f} = t^{-1}\sum_i^t f_i$,
we consider three other constant values: using the best and worst seen 
observation's value at each iteration,
\ie $c = f_{min} = \fstar  = \min \{f_1, f_2, \dots, f_t \}$
and $c = f_{max} = \max \{f_1, f_2, \dots, f_t \}$, and the median observed 
value $c = f_{med} = \median\{f_1, f_2, \dots, f_t \}$. In comparison to using
the data mean $c = \bar{f}$, using $c = f_{min}$ 
leads to acquisition functions becoming more exploratory, as illustrated in 
Figure~\ref{fig:mean_func_example} (left panel). This is because locations far away from 
previously evaluated solutions $\Data$ will have  predicted means
$\mu(\bx)$  equal to 
the $\fstar$ and with large predicted uncertainty, leading to 
large values of, for example, EI and UCB. Conversely, using $c = f_{max}$ will,
as illustrated in Figure~\ref{fig:mean_func_example} (right panel),  lead to increased exploitation
due to regions far from $\Data$ having large uncertainty, but poor
predicted values and hence small $\alpha(\bx)$. Interestingly, 
the effect of using $c = \bar{f}$ or $c = f_{med}$ will
change over the course of the optimisation.
When there are relatively 
few function evaluations $\bar{f}$ and $f_{med}$ will be approximately
$(f_{min} + f_{max})/2$.  However, as the number of expensive function
evaluations increases and the optimisation converges towards the
(estimated) optimum, $\bar{f}$ and particularly $f_{med}$ will tend to approach
$f_{min}$, thus leading to increased exploration.

We also consider linear and quadratic mean functions. Linear mean functions
are defined as $\bhh(\bx) = [1, x^{(1)}, x^{(2)}, \dots, x^{(d)}]$, with 
corresponding weights $\bww \in \Real^{d+1}$ and where $x^{(i)}$ refers to the
$i$th element of  $\bx$. Quadratic mean functions are defined 
similarly, with polynomial terms up to degree $2$ and 
with weights $\bww \in \Real^q$, where $q = \binom{d+2}{d}$. To avoid
overfitting to the data, these regressions are typically trained via a
regularised least-squares approximation, \ie ridge regression (also known as
Tikhonov regularization). The optimal regularised weights $\bww^*$ are
estimated by solving
\begin{equation}
\label{eqn:tikhonov}
\bww^* = \argmin_{\bww} \, \lVert \bff - H \bww \rVert^2 + \lambda \lVert \bww \rVert^2,
\end{equation}
where $H = [\bhh(\bx_1), \bhh(\bx_2), \dots, \bhh(\bx_t) ]$ and
$\lambda \geq 0$ controls the amount of regularisation. The ordinary least 
squares estimator is
\begin{equation}
\label{eqn:tikhonov:soln}
\bww^* = \left(H^\top H + \lambda\bI\right)^{-1} H^\top \bff,
\end{equation}
In this work, the regularisation parameter $\lambda$ was chosen via 
five-fold cross-validation for
$\lambda \in \{10^{-6}, 10^{-5}, \dots, 10^1, 10^2\}$.

Another choice of basis functions are radial basis functions (RBFs), which have
the property that each basis function only depends on the Euclidean distance
from a fixed centre \citep{bishop:prml}. These are known as RBF networks and
can be thought of as either linear neural networks using RBF activation
functions \citep{orr:rbfnetworks} or as finite-dimensional Gaussian processes
\citep{bishop:prml}. A commonly used set of basis functions, and the ones used
in this work, are the Gaussian RBFs
$\phi_i(\bx) = \exp( -\gamma \lVert \bx - \bz_i \rVert )$. While any set of
locations can be used as the centres $\bz_i$, we place a Gaussian RBF at each of
the previously-evaluated locations, \ie $\bz_i \equiv \bx_i ~\forall i = 1,
\ldots, t$, resulting in
$\bhh(\bx) = [\phi_1(\bx), \phi_2(\bx), \dots, \phi_t(\bx)]$. 
Similarly to the
linear and quadratic mean functions, the regularisation parameter $\lambda$ and 
length scale $\gamma$ were chosen via five-fold cross validation. Values of
$\lambda$ were selected in the same range as for the linear and quadratic mean
functions, and the values of $\gamma$ were selected from
$\gamma \in \{ 10^{-3}, 10^{-2.5}, \dots, 10^{1.5}, 10^{2}\}$. A more
fine-grained selection of values were chosen following a preliminary
investigation which revealed the modelling error to be more sensitive to
changes in $\gamma$ than $\lambda$.
We note here that the use of regularisation is
particularly important when placing an RBF on each evaluated location because
the RBF network will otherwise be able to perfectly interpolate the data. This
would lead to $\bff \equiv \bmm$ and therefore \eqref{eqn:mll} would reduce to
$-\tfrac{1}{2} \log \lvert K \rvert$, which can be maximised by either
$\theta_0 \rightarrow 0$ or $\theta_1 \rightarrow \infty$ in \eqref{eqn:mat52}.
This results in the posterior variance estimates  $\sigma^2(\bx)$ being
over-confident and thus having small variance everywhere with predictions
determined by the mean function.

Lastly, we include a non-parametric regressor, extremely randomised trees,
better known as Extra-Trees (ET, \citep{geurts:et}), a variant of Random
Forests (RF, \citep{breiman:rf}). RFs are ensembles of classification or regression trees that are
each trained on a different randomly chosen subsets of the data. Unlike RFs,
that attempt to split the data at each cut-point of a tree optimally, the ET method
instead selects the cut-point as the best from a small set of randomly
chosen cut-points; this additional randomisation results in a
smoother regression in comparison to RFs. Given that ETs use randomised
cut-points, they typically use all the training data in each tree. However, to
counteract the overfitting that this will produce, we allow each item in the
training set to be resampled, instead of using all elements of the set for each
tree. Note that, while not presented as such here, RFs can also be interpreted
as a kernel method \citep{scornet:rfkernels, davies:rfkernels}.

\section{Experimental Evaluation}
\label{sec:results}
We now investigate the performance of the mean functions discussed in 
Section~\ref{sec:meanfuncs} using the EI and UCB acquisition functions
(Section~\ref{sec:bo}) on ten well-known benchmark functions with a range
dimensionality and landscape properties, and two real-world applications. Full 
results of all experimental evaluations are available in the supplementary
material. The mean functions to be evaluated are the constant functions,
$c = \bar{f}$,
$c = f_{med}$, $c = f_{min}$, and $c = f_{max}$, labelled \textit{Arithmetic},
\textit{Median}, \textit{Min}, and \textit{Max} respectively, as well as the
\textit{Linear}, \textit{Quadratic}, Extra-Trees (\textit{RandomForest}), and
RBF network-based (\textit{RBF}) mean functions. 

A Gaussian process surrogate model with an isotropic \Matern $5/2$ kernel
\eqref{eqn:mat52} was
used in all experiments. The Bayesian optimisation runs themselves were
carried out as in Algorithm~\ref{alg:bo}, with the additional step of fitting
a mean function before training the GP
(Algorithm~\ref{alg:bo}, line~\ref{alg:bo:train}) at each iteration. All test 
problems evaluated in this work were scaled to $[0, 1]^d$, and observations
were standardised at each BO iteration, prior to mean function fitting. The 
models were initially trained on $M = 2 d$ observations generated by maximin Latin
hypercube sampling \citep{mckay:lhs}, and each optimisation run was repeated
$51$ times with different initialisation. The same set of $51$ initial
observations were used for each of the mean functions to enable statistical
comparisons. The hyperparameters $\btheta$ of the GP were optimised by
maximising the marginal log likelihood \eqref{eqn:mll} with L-BFGS-B
\citep{byrd:lbfgs} using 10 restarts. Following common practise, maximisation
of the acquisition functions was carried out via multi-start optimisation; 
details of the full procedure can be found in \citep{botorch}. The trade-off 
between exploitation and exploration in UCB, $\beta_t$, is set to Theorem~1
in \citep{srinivas:ucb}, which increases logarithmically with the number of
function evaluations, with $\sqrt{\beta_t}$ approximately in the range
$[3, 6]$. The Bayesian optimisation pipeline and mean functions were
implemented with BoTorch \citep{botorch} and code is available
online\footnote{\url{http://www.github.com/georgedeath/bomean}} to recreate all
experiments, as well as the LHS initialisations used and full optimisation
runs.

Optimisation quality is measured with simple regret $R_t$, which is the 
difference between the true minimum value $f(\bx^*)$ and the best value 
found so far after $t$ evaluations: 
\begin{equation}
R_t = | f(\bx^*) - \min \{f_1, f_2, \dots, f_t \} |.
\end{equation}

\subsection{Synthetic Experiments}
\label{sec:results:synthetic}
\begin{table}[t]
\centering
\begin{tabular}[t]{lr p{0.1\columnwidth} lr}
\addlinespace[-\aboverulesep]\cmidrule[\heavyrulewidth]{1-2}\cmidrule[\heavyrulewidth]{4-5}
    \textbf{Name}  & $d$ && \textbf{Name}  & $d$  \\
\cmidrule[\lightrulewidth]{1-2}\cmidrule[\lightrulewidth]{4-5}
	Branin         & 2   && Ackley         & 5    \\
	Eggholder      & 2   && Hartmann6      & 6    \\
	GoldsteinPrice & 2   && Michalewicz    & 10   \\
	SixHumpCamel   & 2   && Rosenbrock     & 10   \\
	Shekel         & 4   && StyblinskiTang & 10   \\
\cmidrule[\heavyrulewidth]{1-2}\cmidrule[\heavyrulewidth]{4-5}\addlinespace[-\belowrulesep]
\end{tabular}
\caption{Synthetic functions used and their dimensionality $d$. Formulae for 
all functions can be found at
\url{http://www.sfu.ca/~ssurjano/optimization.html}.}
\label{tbl:function_details}
\end{table}
The mean functions were evaluated on the ten popular synthetic benchmark 
functions listed in Table~\ref{tbl:function_details} with a budget of $200$ function
evaluations that included the initial $2d$ LHS samples. These functions were
selected due to their different dimensionality and landscape properties, such
the presence of multiple local or global minima (Ackley, Eggholder, Hartmann6,
GoldsteinPrice, StyblinskiTang, Shekel) deep, valley-like regions (Branin, 
Rosenbrock, SixHumpCamel), and steep ridges and drops (Michalewicz).

  \begin{table*}[t]
  \setlength{\tabcolsep}{2pt}
  \caption{Mean function performance using the EI acquisition function.
           Median regret (\textit{column left}) and median absolute deviation from the
           median (MAD, \textit{column right}) after 200 function evaluations across 
           the 51 runs. The method with the lowest median performance is shown
           in dark grey, with those statistically equivalent shown in light
           grey.}
  \resizebox{1\textwidth}{!}{%
  \begin{tabular}{l Sv Sv Sv Sv Sv}
    \toprule
    \bfseries Mean function
    & \multicolumn{2}{c}{\bfseries Branin (2)} 
    & \multicolumn{2}{c}{\bfseries Eggholder (2)} 
    & \multicolumn{2}{c}{\bfseries GoldsteinPrice (2)} 
    & \multicolumn{2}{c}{\bfseries SixHumpCamel (2)} 
    & \multicolumn{2}{c}{\bfseries Shekel (4)} \\ 
    & \multicolumn{1}{c}{Median} & \multicolumn{1}{c}{MAD}
    & \multicolumn{1}{c}{Median} & \multicolumn{1}{c}{MAD}
    & \multicolumn{1}{c}{Median} & \multicolumn{1}{c}{MAD}
    & \multicolumn{1}{c}{Median} & \multicolumn{1}{c}{MAD}
    & \multicolumn{1}{c}{Median} & \multicolumn{1}{c}{MAD}  \\ \midrule
    Arithmetic & 1.35e-05 & 1.82e-05 & \best 1.58e+00 & \best 1.93e+00 & \best 4.18e-02 & \best 5.32e-02 & 2.51e-05 & 2.58e-05 & \statsimilar 8.13e-02 & \statsimilar 1.19e-01 \\
    Median & 9.35e-06 & 1.17e-05 & \statsimilar 3.02e+00 & \statsimilar 2.67e+00 & \statsimilar 5.03e-02 & \statsimilar 5.59e-02 & 1.40e-05 & 1.49e-05 & \statsimilar 1.54e-01 & \statsimilar 2.24e-01 \\
    Min & 1.18e-05 & 1.31e-05 & \statsimilar 2.82e+00 & \statsimilar 3.13e+00 & \statsimilar 6.25e-02 & \statsimilar 8.03e-02 & 2.25e-05 & 2.29e-05 & 7.02e+00 & 9.75e-01 \\
    Max & 7.21e-06 & 8.81e-06 & \statsimilar 2.69e+00 & \statsimilar 2.48e+00 & \statsimilar 6.87e-02 & \statsimilar 7.40e-02 & 1.52e-05 & 1.84e-05 & \best 7.16e-02 & \best 1.05e-01 \\
    Linear & \best 5.13e-06 & \best 5.58e-06 & \statsimilar 2.82e+00 & \statsimilar 2.85e+00 & \statsimilar 1.05e-01 & \statsimilar 1.19e-01 & \statsimilar 7.98e-06 & \statsimilar 8.14e-06 & 6.47e+00 & 1.57e+00 \\
    Quadratic & 1.05e-05 & 1.09e-05 & \statsimilar 3.59e+00 & \statsimilar 4.03e+00 & \statsimilar 5.55e-02 & \statsimilar 6.03e-02 & \best 6.27e-06 & \best 6.52e-06 & 6.47e+00 & 1.30e+00 \\
    RandomForest & 5.35e-04 & 4.86e-04 & \statsimilar 3.93e+00 & \statsimilar 4.22e+00 & 2.41e+00 & 2.14e+00 & 2.96e-04 & 3.56e-04 & 2.84e+00 & 2.76e+00 \\
    RBF & 1.36e-04 & 1.07e-04 & \statsimilar 3.27e+00 & \statsimilar 3.45e+00 & 1.76e+00 & 2.39e+00 & 3.76e-05 & 4.63e-05 & 7.95e+00 & 1.05e+00 \\
\bottomrule
    \toprule
    \bfseries Mean function
    & \multicolumn{2}{c}{\bfseries Ackley (5)} 
    & \multicolumn{2}{c}{\bfseries Hartmann6 (6)} 
    & \multicolumn{2}{c}{\bfseries Michalewicz (10)} 
    & \multicolumn{2}{c}{\bfseries Rosenbrock (10)} 
    & \multicolumn{2}{c}{\bfseries StyblinskiTang (10)} \\ 
    & \multicolumn{1}{c}{Median} & \multicolumn{1}{c}{MAD}
    & \multicolumn{1}{c}{Median} & \multicolumn{1}{c}{MAD}
    & \multicolumn{1}{c}{Median} & \multicolumn{1}{c}{MAD}
    & \multicolumn{1}{c}{Median} & \multicolumn{1}{c}{MAD}
    & \multicolumn{1}{c}{Median} & \multicolumn{1}{c}{MAD}  \\ \midrule
    Arithmetic & 4.27e+00 & 6.06e+00 & 4.00e-03 & 5.46e-03 & 7.22e-02 & 1.03e-01 & \statsimilar 8.38e+02 & \statsimilar 3.27e+02 & 6.47e+01 & 2.58e+01 \\
    Median & 2.10e+00 & 2.06e+00 & 8.31e-04 & 1.02e-03 & 7.88e-02 & 1.12e-01 & \statsimilar 7.14e+02 & \statsimilar 2.78e+02 & 6.72e+01 & 2.49e+01 \\
    Min & 4.64e+00 & 9.59e-01 & 3.27e-03 & 3.72e-03 & 1.39e+00 & 7.49e-01 & \statsimilar 7.00e+02 & \statsimilar 2.21e+02 & 8.15e+01 & 2.78e+01 \\
    Max & \best 1.66e+00 & \best 1.23e+00 & \best 7.47e-04 & \best 9.88e-04 & \best 2.75e-02 & \best 3.72e-02 & \best 6.95e+02 & \best 3.64e+02 & \best 2.84e+01 & \best 2.08e+01 \\
    Linear & 1.65e+01 & 2.42e+00 & 1.21e-02 & 1.32e-02 & 1.30e+00 & 8.55e-01 & 1.93e+03 & 1.02e+03 & 1.10e+02 & 2.99e+01 \\
    Quadratic & 9.83e+00 & 5.81e+00 & 9.95e-03 & 9.63e-03 & 1.03e+00 & 6.84e-01 & 1.81e+03 & 8.90e+02 & 8.58e+01 & 4.14e+01 \\
    RandomForest & 5.16e+00 & 7.70e-01 & 8.12e-02 & 5.63e-02 & 1.25e-01 & 1.43e-01 & 4.21e+03 & 2.29e+03 & 7.27e+01 & 1.78e+01 \\
    RBF & 7.90e+00 & 1.94e+00 & 2.75e-01 & 1.55e-01 & 9.05e-01 & 5.57e-01 & 2.73e+03 & 1.32e+03 & 1.17e+02 & 2.86e+01 \\
\bottomrule
  \end{tabular}
  }
  \label{tbl:synthetic_results:ei}
  \end{table*}

Table~\ref{tbl:synthetic_results:ei} shows the median regret over the 51
repeated experiments, together with the median absolute deviation from the
median (MAD) for the mean functions using the EI acquisition function. Due to
space constraints, the corresponding table for UCB is included in the supplementary material.
The method with the lowest (best) median regret on each function is highlighted
in dark grey, and those highlighted in light grey are statistically equivalent
to the best method according to a one-sided paired Wilcoxon signed-rank test
\citep{knowles:testing} with Holm-Bonferroni correction \citep{holm:test}
($p\geq0.05$).

\begin{figure*}[t]
\rule{0mm}{4mm}\\  %
\centering
\includegraphics[width=\textwidth, clip, trim={0 0 0 0}]{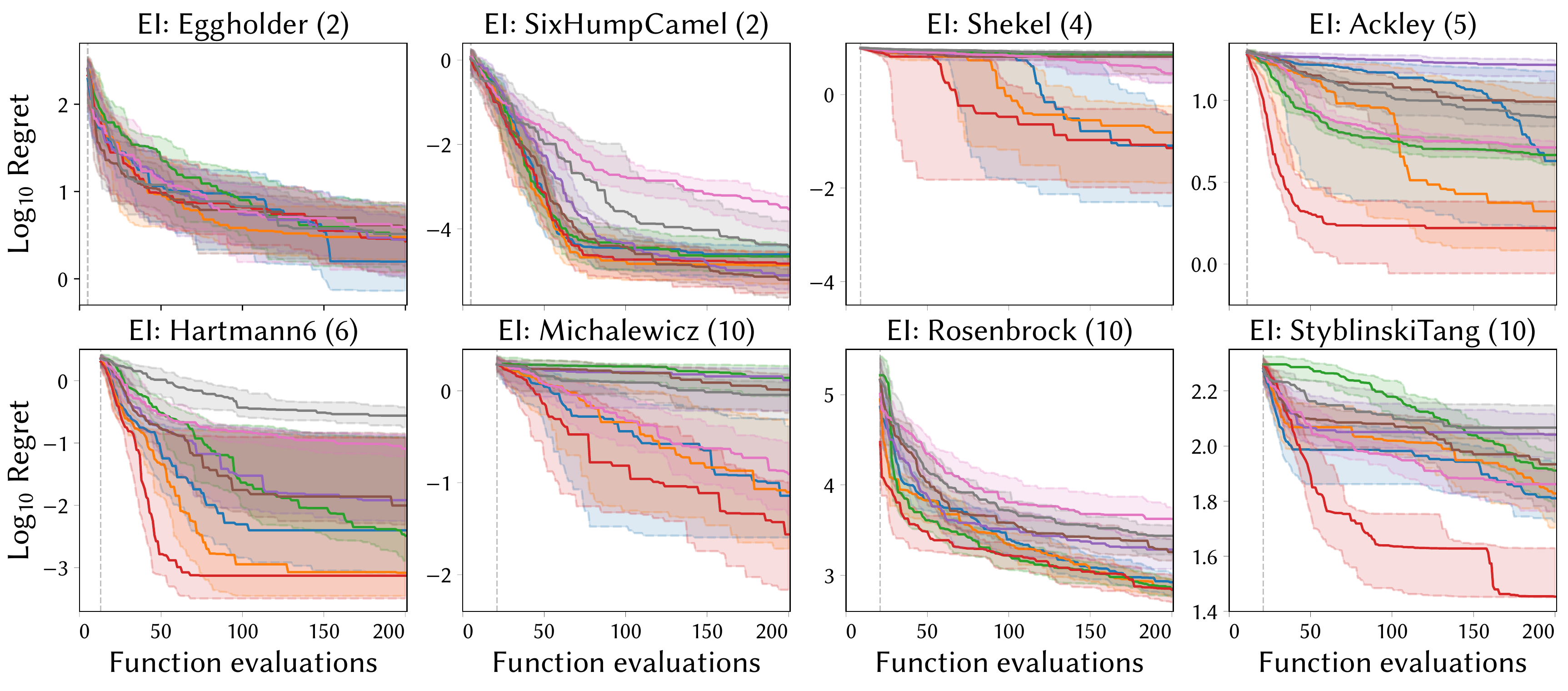}\\
\includegraphics[width=\textwidth, clip, trim={0 10 0 10}]{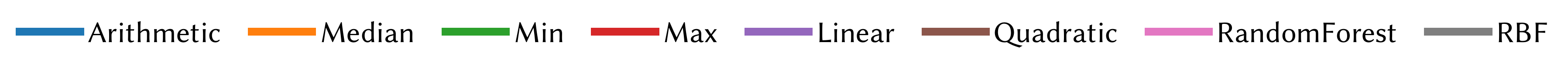}%
\caption{Illustrative convergence plots for eight benchmark problems using the
         EI %
         acquisition function. Each plot shows the median regret, with shading 
         representing the interquartile range across the 51 runs and the dashed
         vertical line indicating the end of the initial LHS phase.}
\label{fig:conv_plots_ei}
\end{figure*}

\begin{figure*}[t]
\centering
\includegraphics[width=\textwidth, clip, trim={0 0 0 0}]{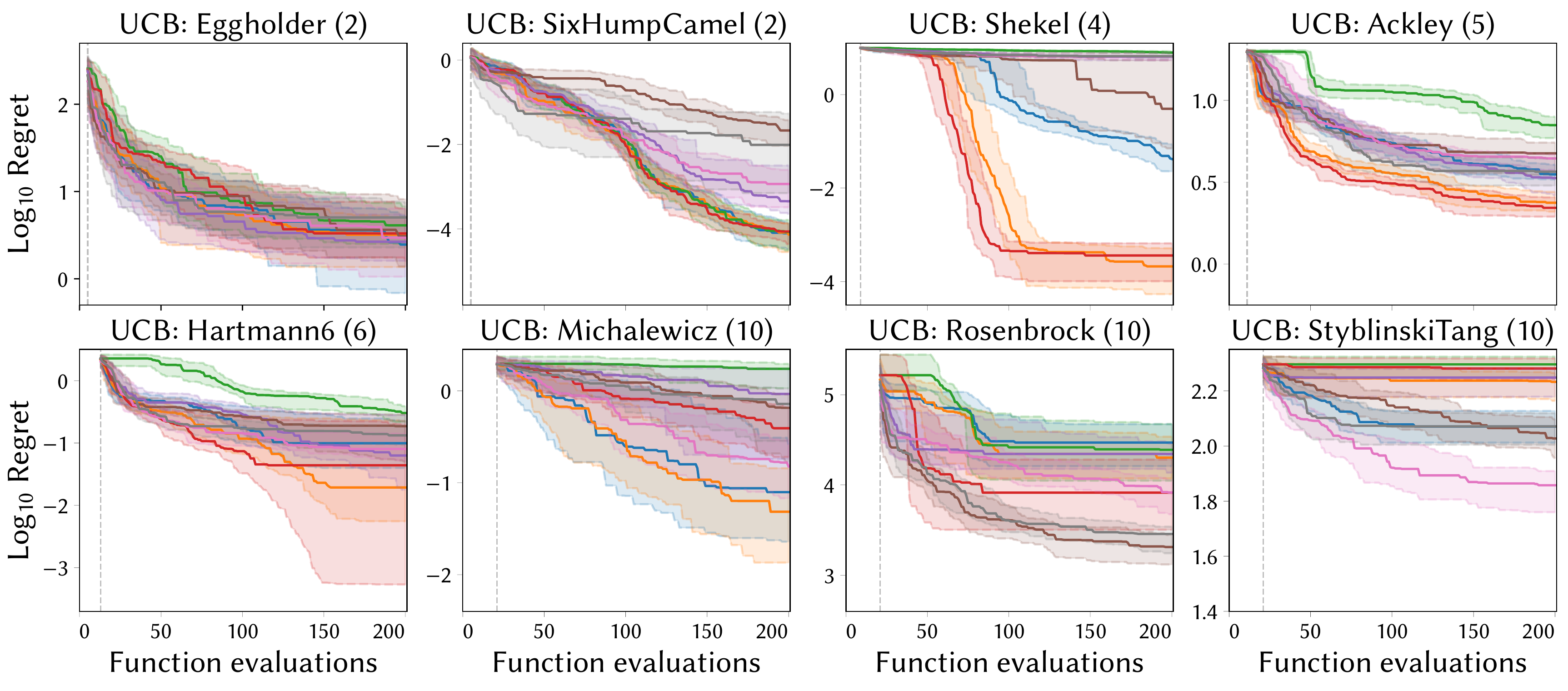}\\
\includegraphics[width=\textwidth, clip, trim={0 10 0 10}]{figs/legend_twocol}%
\caption{Illustrative convergence plots for eight benchmark problems using the
         UCB %
         acquisition function. Each plot shows the median regret, with shading 
         representing the interquartile range across the 51 runs and the dashed
         vertical line indicating the end of the initial LHS phase.}
\label{fig:conv_plots_ucb}
\end{figure*}

The convergence of the various mean functions 
on $8$ illustrative test problems are shown using the EI (Figure~\ref{fig:conv_plots_ei}) and UCB (Figure~\ref{fig:conv_plots_ucb})
acquisition functions. Convergence plots for the Branin and GoldsteinPrice
test problems were visually similar to Eggholder and SixHumpCamel respectively;
they are available in the supplementary material. As one might expect, because
points are naturally less distant from one another, the choice of mean
function has less impact in 2 dimensions. Although, interestingly, optimisation
runs with the UCB algorithm in $d=2$ achieve lower regret with the constant
mean functions compared to the others evaluated.

Perhaps surprisingly, the non-constant mean functions, \textit{Linear},
\textit{Quadratic}, \textit{RandomForest} and \textit{RBF}, appear to offer
no advantage over the constant mean functions despite their ability to
model the large scale optimisation landscape.  The \textit{Quadratic}
model, in two dimensions where there are only three parameters to be fitted,
appears to be well suited to the SixHumpCamel function which is roughly
bowl-shaped (albeit with quartic terms); however, this appears to be an
exceptional case. 

In higher dimensions with the EI acquisition function, using the worst 
observation value as the constant mean function (\textit{Max}) consistently 
provides the lowest regret on the test functions evaluated. This is
consistent with recent
work \citep{death:egreedy, rehbach:ei_pv} showing that being more exploitative
in higher dimensions is preferable to most other strategies. However, for the
UCB acquisition function this is not the case and no mean function is
consistently best. We suspect that this is because
the value of $\beta_t$ is so large that the UCB function \eqref{eqn:ucb} is always
dominated by the exploratory term ($\sqrt{\beta_t}\sigma(\bx)$) so that the
mean function has relatively little influence. This is in contrast to EI,
which has been shown
\citep{death:egreedy} to be far more exploitative than UCB.

The standard choice of using a  constant mean function equal to $\bar{f}$ the
arithmetic mean of the observations  (\textit{Arithmetic}) is, in
higher dimensions ($d \geq 5$),
only statistically equivalent to the best-performing method on one of the five
test functions for both EI and UCB. This result calls into question the
efficacy of the common practise of using
the $\bar{f}$ constant mean function in all Bayesian optimisation tasks. Based
on these results we suggest that an increase performance may be obtained by using
the \textit{Max} mean function with EI. Although there does not appear to be
such a clear-cut answer as to which mean function should be used in conjunction
with the UCB acquisition function, we posit that this is less important
because, based on these optimisation results, one would prefer the performance
of EI over UCB in general.

\subsection{Active Learning for Robot Pushing}
\label{sec:results:robots}
Like \citep{wang:MES, jiang:nonmyopicbo, death:egreedy, death:eshotgun} we
optimise the control parameters for two active learning robot pushing problems
\citep{wang:robots}; see \citep{death:egreedy} for a diagrammatic outline of
the problems. In the $d = 4$ \pushfour problem, a robot should push
an object towards an unknown target location and is constrained such that it
can only travel in the initial direction of the object. Once the robot has
finished pushing, it receives feedback in the form of the object-target
distance. The robot is parametrised by its initial location, the orientation
of its pushing hand and how long it pushes for. This can therefore be cast as a
minimisation problem in which the four parameter values are optimised with
respect to the object's final distance from the target. The object's initial
location is fixed to the centre of the problem domain \citep{wang:MES} and the
target location is changed for each of the 51 optimisation runs, with these
kept the same across mean functions. Thus, the optimisation performance is 
considered over problem instances rather initialisations of a single instance.

The second problem, \pusheight, two robots push their respective objects
towards two unknown targets, with their movements constrained so that they
always travel in the direction of their object's initial location. The $d = 8$
parameters controlling the robots can be optimised to minimise the summed final
object-target distances. Like \pushfour, initial object locations were fixed
for problem instances and the targets' locations were chosen randomly, with a
constraint enforcing that both objects could cover the targets without 
overlapping. This means, however, that in some problem instances it may not be
possible for both robots to push their objects to their respective targets 
because they will block each other. Thus, for \pusheight we report the final
summed object-distances rather than the regret due to the global optimum not
being known.

\begin{figure}[t]
\centering
\includegraphics[width=\columnwidth, clip, trim={0 0 0 0}]{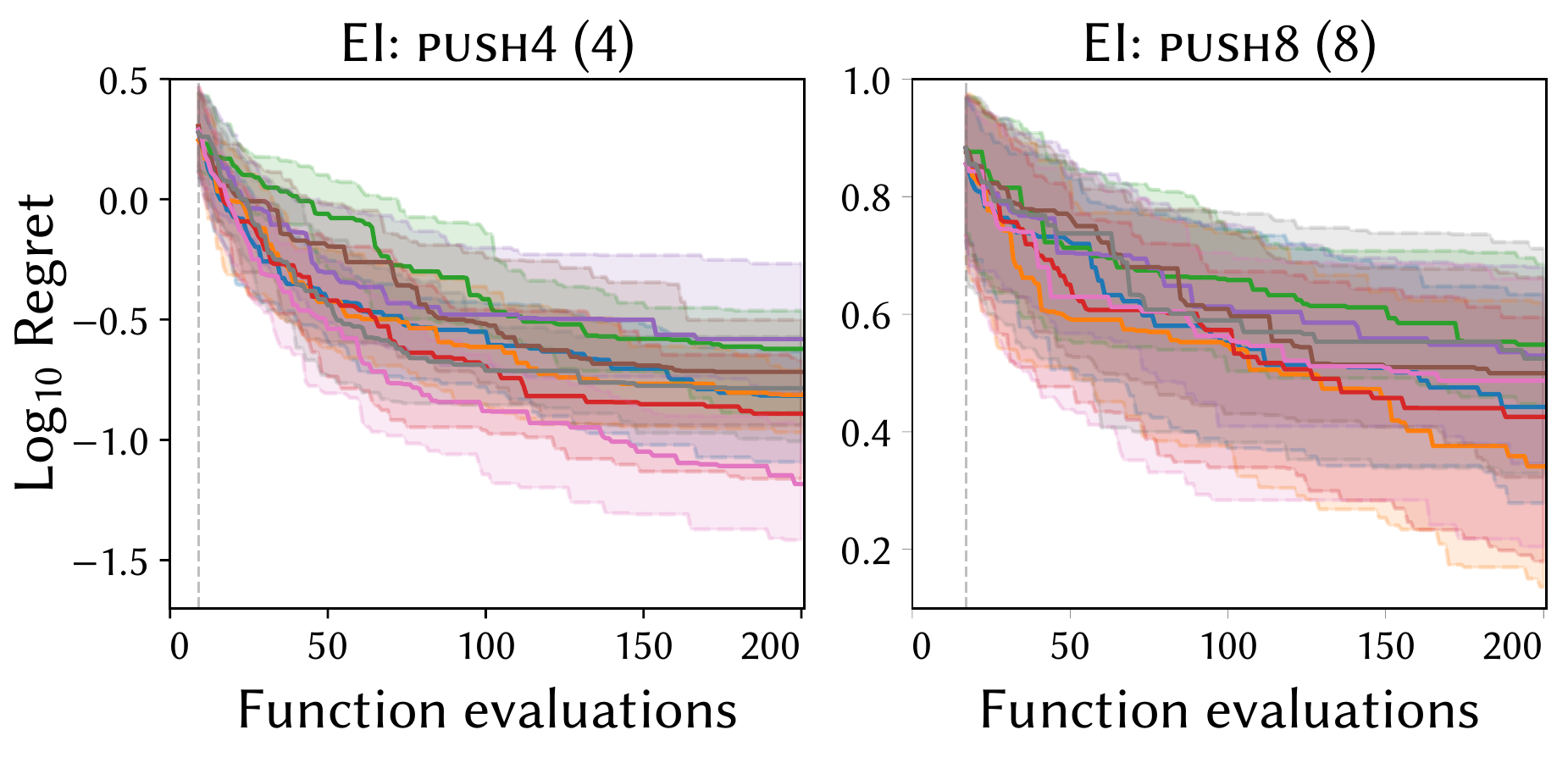}\\
\includegraphics[width=\columnwidth, clip, trim={0 0 0 0}]{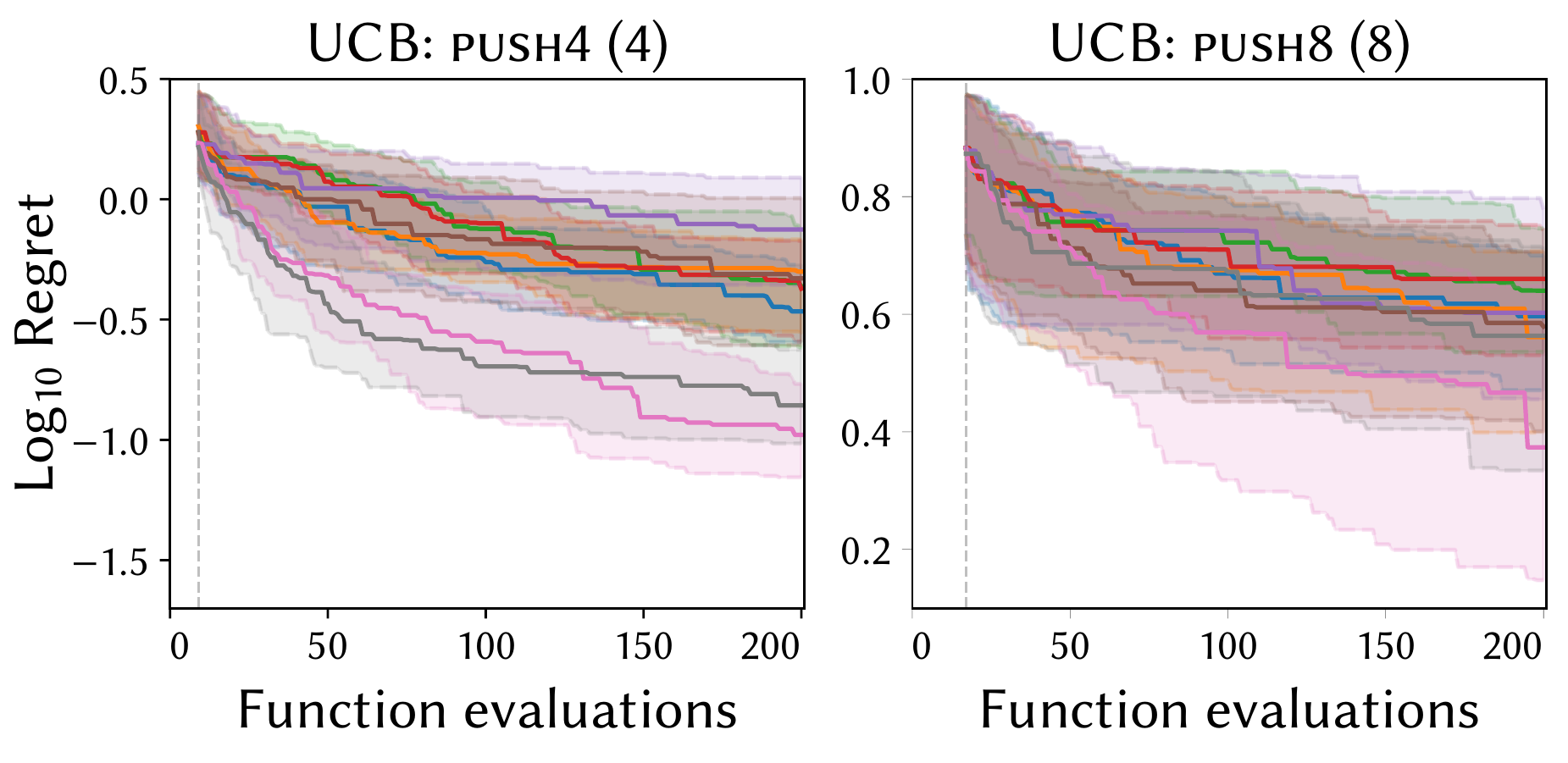}\\
\includegraphics[width=\columnwidth, clip, trim={0 10 0 10}]{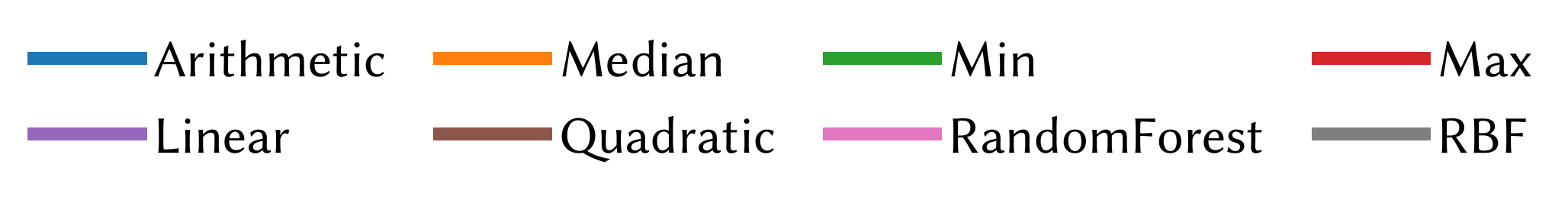}%
\caption{Convergence plots for the robot pushing problem using EI and UCB. Each
         plot shows the median regret, with shading representing the
         interquartile range of the 51 runs.}
\label{fig:conv_plots_push}
\end{figure}
Figure~\ref{fig:conv_plots_push} shows the convergence plots using the various
mean functions with the EI and UCB acquisition functions. Tabulated results
are available in the supplementary material. In the four-dimensional \pushfour
problem, one of the worst-performing mean functions on the synthetic functions,
\textit{RandomForest}, had substantially lower regret than the
other mean functions using UCB on both problems and for EI on \pushfour. However,
in the eight-dimensional \pusheight, 
all mean functions were statistically equivalent when using EI, apart from
\textit{Min}, \textit{Linear} and \textit{RBF}, which were worse. However, when
using UCB, the \textit{RandomForest} mean function was statistically better 
than all other mean functions, and it achieved results comparable to using EI.

It might be suspected that the superior  ability of the Random
Forest (and RBF on some problems) to represent the inherently difficult
landscape features of these problems would account for the better performance
of \textit{RandomForest}. The landscape for these problems has
sharp changes (high gradients) in the  function,
\eg when a change of the robot's starting location results in the object no 
longer being
pushed towards the target, as well as plateaux where changes in certain
parameter values have little  effect, \eg if the amount of pushing time
results in the robot not reaching the object.
However, we investigated the mean prediction
error of each of the combined mean plus Gaussian process models trained on
the first 100 expensively evaluated locations by calculating the normalised
root mean squared prediction error at 1000 locations (chosen by Latin
Hypercube Sampling). As shown in the supplementary material, this indicates
that, in fact, the RF model has a comparatively poor prediction error. By
contrast, the RBF mean function yields the most accurate model of the
overall landscape, but it only shows better performance for \pushfour using
UCB. These prediction errors were evaluated over the entire domain and,
therefore, it is possible that the RF mean function is sufficiently superior
in the vicinity of the optimum to allow the more rapid convergence seen here.

Interestingly, the performance of the \textit{Max} mean function with EI on the
synthetic test problems is not reflected on these two more real-world problems.
However, both the \textit{Arithmetic} and \textit{Max} mean functions are
statistically equivalent on both \pushfour and \pusheight, with \textit{Max}
having lower median regret and MAD than \textit{Arithmetic}.
Nonetheless, it  remains unclear why the RF mean function gives best
performance in three of these four cases.

\subsection{Pipe Shape Optimisation}
\label{sec:results:pitzdaily}
Lastly, we evaluate the mean functions on a real-world computational fluid
dynamics (CFD) optimisation problem. The goal of the PitzDaily CFD problem
\citep{daniels:benchmark} is to minimise the pressure loss between a pipe's
entrance (inflow) and exit (outflow) by optimising the shape of the pipe's 
lower wall. The loss is evaluated by generating a CFD mesh and simulating the
two-dimensional flow using OpenFOAM \citep{jasak:openfoam}, with each function
evaluation taking between $60$ and $90$ seconds. The decision variables in the
problem are the control points of a Catmull-Clark subdivision curve that 
represents the lower wall's shape; see \citep{daniels:benchmark} for a 
pictorial representation of these. In this work we use $5$ control points,
resulting in a 10-dimensional decision vector. The control points are
constrained to lie in a polygon, rather than a hypercube for all previous 
problems, and, therefore, we draw initial samples uniformly from within the
constrained region rather than using LHS. Similarly, we use CMA-ES
\citep{hansen:bipop} to optimise the acquisition functions and penalise
locations that violate the constraints.

\begin{figure}[t]
\centering%
\includegraphics[width=\columnwidth, clip, trim={0 0 0 0}]{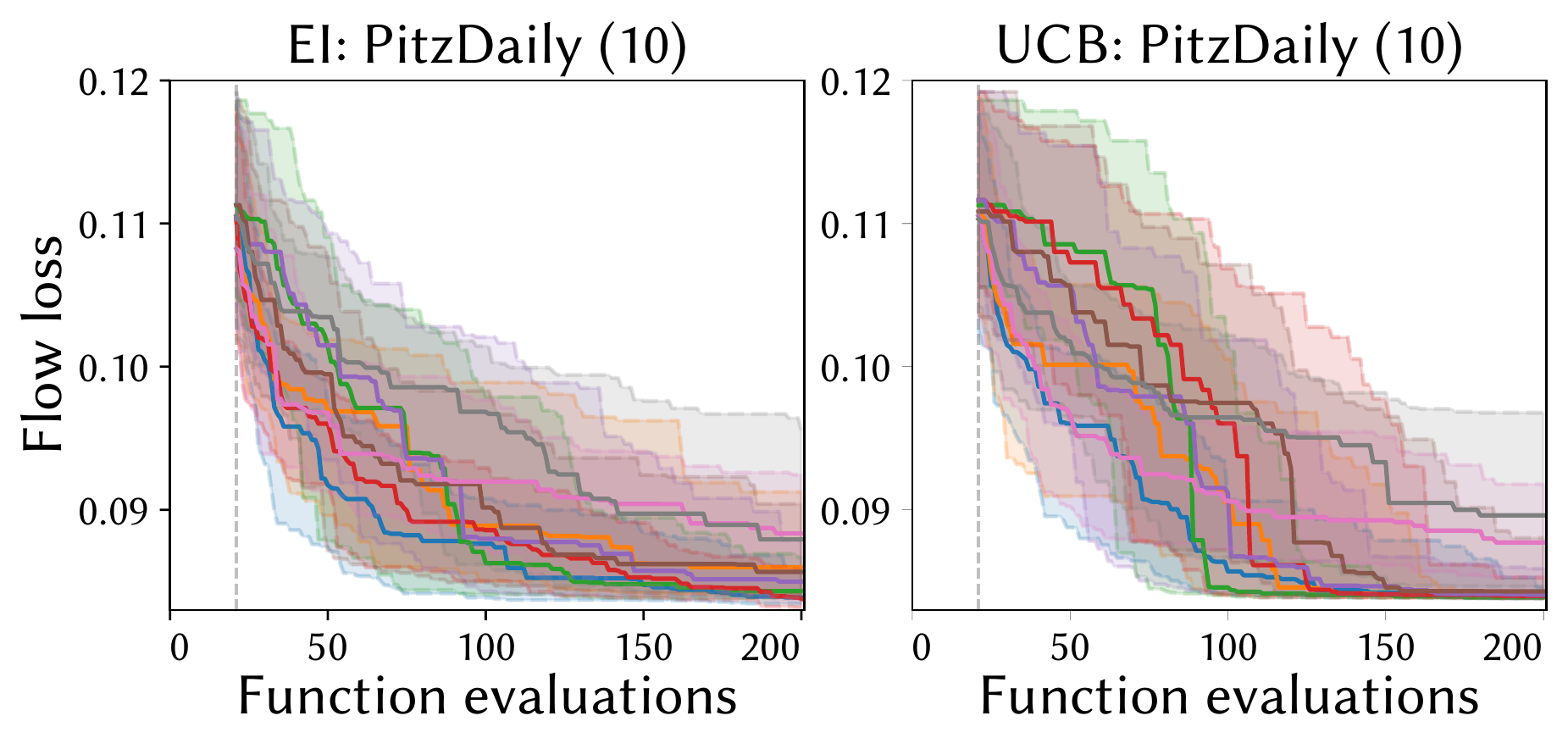}\\
\includegraphics[width=\columnwidth, clip, trim={0 10 0 10}]{figs/legend_onecol}%
\caption{Convergence plots for the PitzDaily test problem using EI and UCB.
         Each plot shows the median regret, with shading representing the
         interquartile range of the 51 runs.}
\label{fig:conv_plots_pitz}
\end{figure}
Convergence plots of the flow loss for the mean functions with EI and UCB are
shown in Figure~\ref{fig:conv_plots_pitz}. The \textit{Arithmetic},
\textit{Min} and \textit{Max} constant mean functions, when using EI, are all
statistically equivalent and the best-performing. It is interesting to note
the contrasting performance between the \textit{RandomForest} mean function 
on this and the robot pushing tasks. %

As shown in Figure~\ref{fig:conv_plots_pitz}, using the UCB acquisition
function leads to the constant mean and \textit{Linear} mean functions all 
having a median flow loss within $10^{-4}$ of one another, with their 
inter-quartile ranges rapidly decreasing. This effect can also be seen towards 
the end of the optimisation runs with EI. Inspection of solutions (control 
points) with a flow loss $\approx 0.084$ revealed that they all had distinct 
values but that they led to very similar sub-division curves. This implies that
they all represented essentially the same inner wall shape and thus indicate
the presence of either one large, valley-like global optimum or many, global 
optima. We suggest that this may be the actual minimum flow loss achievable for
this problem. All mean functions, in combination with both EI and UCB, were 
able to successfully discover solutions that led to a flow loss of less than 
$0.0903$ found by a local, gradient-based method in \citep{nilsson:pitzdaily}
that used approximately 500 function evaluations. This highlights the strength
of Bayesian optimisation in general because the convergence rates shown in 
Figure~\ref{fig:conv_plots_pitz} are far more rapid for the majority of mean
functions and realise better solutions than the local, adjoint method.

\section{Conclusion}
\label{sec:conc}
We have investigated the effect of using different prior mean functions in
the Gaussian process model during
Bayesian optimisation when using the expected improvement and upper confidence
bound acquisition functions. This was assessed by performing BO on ten
synthetic functions and two real-world problems. The constant mean function
\textit{Max}, which uses a constant value of the worst-seen expensive function
evaluation thus far, was found to consistently out-perform the other mean
functions across the synthetic functions in higher dimensions, and was
statistically equivalent to the best performing mean function on nine out of 
ten functions.  We suggest that this is because this mean function tends to
promote exploitation which can lead to rapid convergence in higher
dimensions \citep{death:egreedy, rehbach:ei_pv} because exploration is
implicitly provided through the necessarily inaccurate surrogate modelling.
However, on the two, real-world problems this trend did not
continue, but its performance was still statistically equivalent to the
commonly-used mean equal to the arithmetic mean of the observations.
For this reason we recommend using the
\textit{Max} mean function in conjunction with expected improvement for, at
worst, the same performance as a zero mean and generally improved performance
in higher dimensions.

Interestingly, the lack of consistency between mean function performance on the
synthetic and real-world problems may indicate a larger issue in BO, namely
that synthetic benchmarks do not always contain the same types of functional
landscapes as real-world problems. In future work we would like to characterise
a function's landscape during the optimisation procedure and adaptively select
the best-performing components of the BO pipeline, \eg mean function, kernel,
and acquisition function, to suit the problem structure. 

This work focused on learning a mean function independent of the training of
the GP. In further work we would like to jointly learn the parameters of the
mean function, \ie the weights in \eqref{eqn:mf}, alongside the 
hyperparameters of the GP itself.   The efficacy of the BO approach clearly
depends crucially on the ability of the surrogate model to accurately
predict the function's value in unvisited locations.  We therefore look
forward to evaluating a fully-Bayesian approach that marginalises
over the mean function parameters and
kernel hyperparameters.  Although the Monte Carlo sampling required to
evaluate the resulting acquisition functions may be substantial, an
important area of investigation is whether fully-Bayesian models can
significantly improve the convergence of Bayesian Optimisation.

\begin{acks} 
This work was supported by 
\grantsponsor{}{Innovate UK}{} [grant number \grantnum{}{104400}].
\end{acks}

\balance %
\bibliographystyle{ACM-Reference-Format}
\bibliography{ref}

\end{document}


\title[The Role of the Mean Function in Bayesian Optimisation]
{What do you Mean?\\The Role of the Mean Function in Bayesian Optimisation}
\subtitle{Supplementary Material}

\author{George {De Ath}}
\email{g.de.ath@exeter.ac.uk}
\orcid{0000-0003-4909-0257}
\affiliation{%
  \department{Department of Computer Science}
  \institution{University of Exeter}
  \city{Exeter}
  \country{United Kingdom}
}

\author{Jonathan E. Fieldsend}
\email{j.e.fieldsend@exeter.ac.uk}
\orcid{0000-0002-0683-2583}
\affiliation{%
  \department{Department of Computer Science}
  \institution{University of Exeter}
  \city{Exeter}
  \country{United Kingdom}
}

\author{Richard M. Everson}
\email{r.m.everson@exeter.ac.uk}
\orcid{0000-0002-3964-1150}
\affiliation{%
  \department{Department of Computer Science}
  \institution{University of Exeter}
  \city{Exeter}
  \country{United Kingdom}
}

\maketitle
\appendix
\section{Additional Results}
In this section we show convergence plots for the Branin and GoldsteinPrice
synthetic test functions for both the EI and UCB acquisition functions, as well
as the results tables for UCB on the synthetic functions, and EI and UCB on the
three real-world problems. 

\begin{figure}[H]
\centering%
\includegraphics[width=0.5\textwidth, clip, trim={0 0 0 0}]{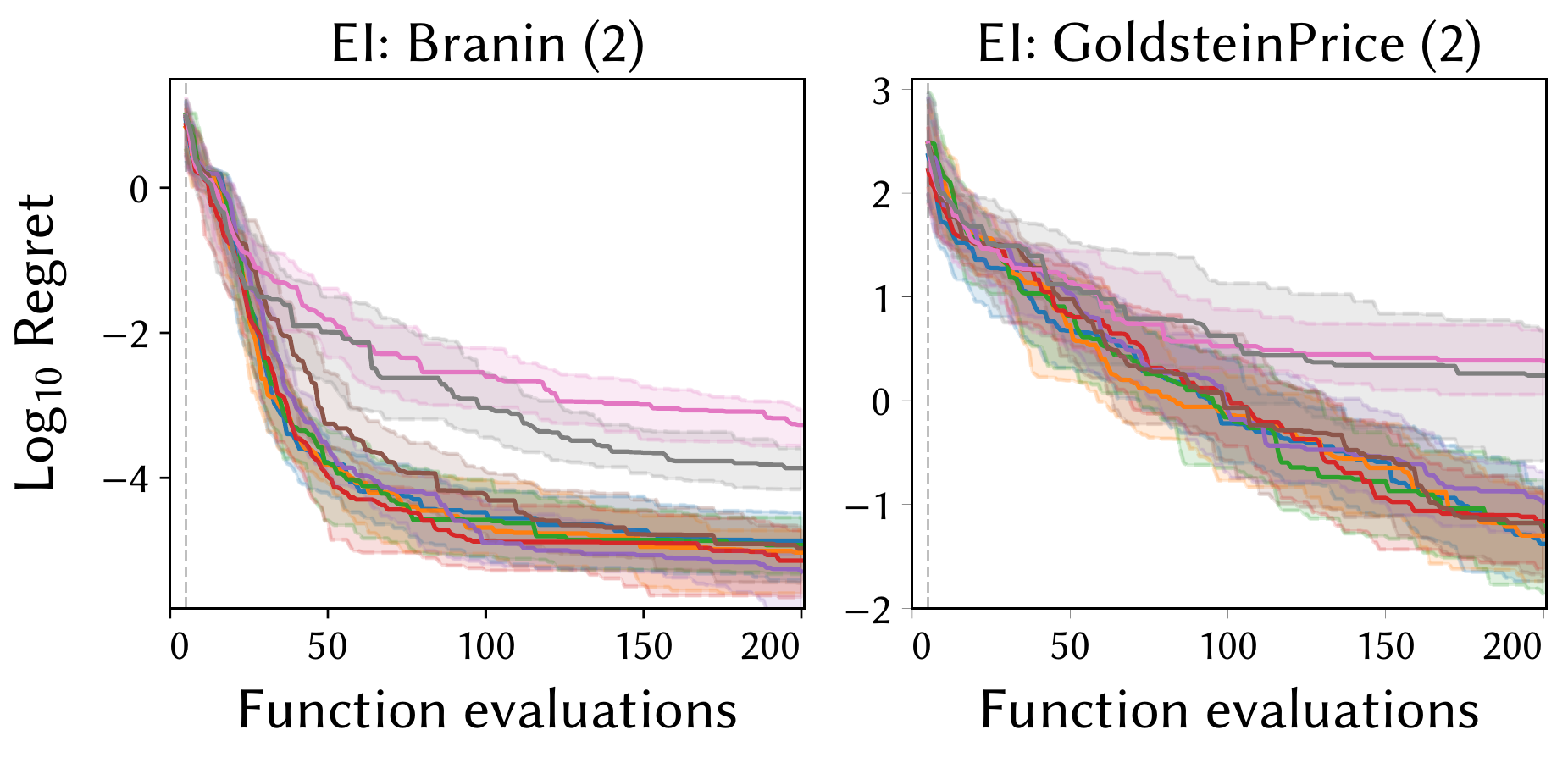}%
\includegraphics[width=0.5\textwidth, clip, trim={0 0 0 0}]{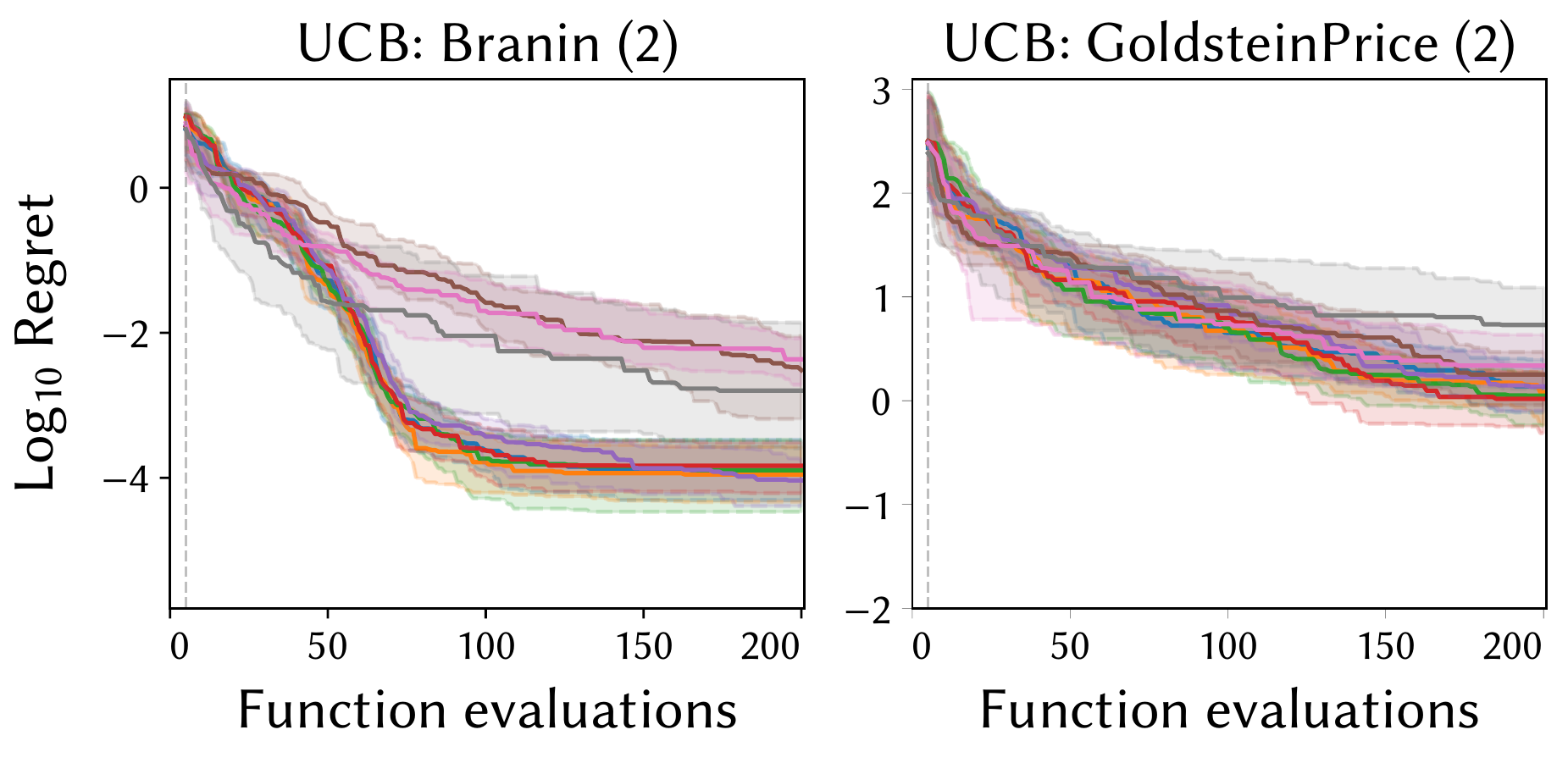}\\
\includegraphics[width=1\textwidth, clip, trim={0 10 0 10}]{figs/legend_twocol}%
\caption{Convergence plots for the Branin and GoldsteinPrice test problem 
         using EI (\textit{left}) and UCB (\textit{right}). Each plot shows the
         median regret, with shading representing the interquartile range
         across the 51 runs.}
\label{fig:conv_plots_supp}
\end{figure}

  \begin{table}[H]
  \setlength{\tabcolsep}{2pt}
  \caption{Mean function performance using the UCB acquisition function.
           Median regret (\textit{column left}) and median absolute deviation from the
           median (MAD, \textit{column right}) after 200 function evaluations across 
           the 51 runs. The method with the lowest median performance is shown
           in dark grey, with those statistically equivalent shown in light
           grey.}
  \resizebox{1\textwidth}{!}{%
  \begin{tabular}{l Sv Sv Sv Sv Sv}
    \toprule
    \bfseries Mean function
    & \multicolumn{2}{c}{\bfseries Branin (2)} 
    & \multicolumn{2}{c}{\bfseries Eggholder (2)} 
    & \multicolumn{2}{c}{\bfseries GoldsteinPrice (2)} 
    & \multicolumn{2}{c}{\bfseries SixHumpCamel (2)} 
    & \multicolumn{2}{c}{\bfseries Shekel (4)} \\ 
    & \multicolumn{1}{c}{Median} & \multicolumn{1}{c}{MAD}
    & \multicolumn{1}{c}{Median} & \multicolumn{1}{c}{MAD}
    & \multicolumn{1}{c}{Median} & \multicolumn{1}{c}{MAD}
    & \multicolumn{1}{c}{Median} & \multicolumn{1}{c}{MAD}
    & \multicolumn{1}{c}{Median} & \multicolumn{1}{c}{MAD}  \\ \midrule
    Arithmetic & \statsimilar 1.27e-04 & \statsimilar 1.47e-04 & \best 2.47e+00 & \best 2.77e+00 & \statsimilar 1.37e+00 & \statsimilar 1.24e+00 & \statsimilar 7.94e-05 & \statsimilar 7.12e-05 & 4.20e-02 & 3.38e-02 \\
    Median & \statsimilar 1.11e-04 & \statsimilar 1.44e-04 & \statsimilar 3.21e+00 & \statsimilar 2.76e+00 & \statsimilar 1.21e+00 & \statsimilar 9.88e-01 & \best 7.44e-05 & \best 7.44e-05 & \best 2.11e-04 & \best 2.89e-04 \\
    Min & \statsimilar 1.27e-04 & \statsimilar 1.57e-04 & \statsimilar 4.09e+00 & \statsimilar 3.09e+00 & \statsimilar 1.12e+00 & \statsimilar 9.27e-01 & \statsimilar 8.57e-05 & \statsimilar 9.33e-05 & 7.88e+00 & 6.60e-01 \\
    Max & \statsimilar 1.46e-04 & \statsimilar 1.55e-04 & \statsimilar 3.31e+00 & \statsimilar 3.01e+00 & \best 1.04e+00 & \best 1.09e+00 & \statsimilar 8.75e-05 & \statsimilar 7.29e-05 & \statsimilar 3.63e-04 & \statsimilar 4.03e-04 \\
    Linear & \best 9.20e-05 & \best 8.31e-05 & \statsimilar 2.66e+00 & \statsimilar 2.34e+00 & \statsimilar 1.36e+00 & \statsimilar 8.95e-01 & 4.53e-04 & 5.47e-04 & 6.47e+00 & 7.89e-02 \\
    Quadratic & 3.05e-03 & 4.36e-03 & \statsimilar 3.08e+00 & \statsimilar 4.15e+00 & 1.78e+00 & 1.59e+00 & 2.15e-02 & 2.01e-02 & 4.98e-01 & 7.31e-01 \\
    RandomForest & 4.30e-03 & 4.55e-03 & \statsimilar 2.96e+00 & \statsimilar 3.30e+00 & 2.18e+00 & 1.49e+00 & 1.16e-03 & 1.40e-03 & 6.61e+00 & 1.76e+00 \\
    RBF & 1.59e-03 & 2.21e-03 & 5.05e+00 & 4.05e+00 & 5.36e+00 & 6.16e+00 & 9.78e-03 & 1.43e-02 & 6.64e+00 & 1.60e+00 \\
\bottomrule
    \toprule
    \bfseries Mean function
    & \multicolumn{2}{c}{\bfseries Ackley (5)} 
    & \multicolumn{2}{c}{\bfseries Hartmann6 (6)} 
    & \multicolumn{2}{c}{\bfseries Michalewicz (10)} 
    & \multicolumn{2}{c}{\bfseries Rosenbrock (10)} 
    & \multicolumn{2}{c}{\bfseries StyblinskiTang (10)} \\ 
    & \multicolumn{1}{c}{Median} & \multicolumn{1}{c}{MAD}
    & \multicolumn{1}{c}{Median} & \multicolumn{1}{c}{MAD}
    & \multicolumn{1}{c}{Median} & \multicolumn{1}{c}{MAD}
    & \multicolumn{1}{c}{Median} & \multicolumn{1}{c}{MAD}
    & \multicolumn{1}{c}{Median} & \multicolumn{1}{c}{MAD}  \\ \midrule
    Arithmetic & 3.53e+00 & 6.53e-01 & 9.93e-02 & 1.22e-01 & \statsimilar 7.90e-02 & \statsimilar 9.42e-02 & 2.95e+04 & 2.36e+04 & 1.18e+02 & 2.40e+01 \\
    Median & 2.37e+00 & 5.60e-01 & \best 1.94e-02 & \best 2.70e-02 & \best 4.84e-02 & \best 5.89e-02 & 2.00e+04 & 1.61e+04 & 1.71e+02 & 2.97e+01 \\
    Min & 7.06e+00 & 1.49e+00 & 3.03e-01 & 1.29e-01 & 1.74e+00 & 6.51e-01 & 2.45e+04 & 2.21e+04 & 1.98e+02 & 2.57e+01 \\
    Max & \best 2.21e+00 & \best 4.51e-01 & \statsimilar 4.40e-02 & \statsimilar 6.48e-02 & 3.92e-01 & 3.59e-01 & 8.22e+03 & 8.89e+03 & 1.91e+02 & 2.52e+01 \\
    Linear & 3.35e+00 & 8.75e-01 & 6.31e-02 & 8.03e-02 & 9.27e-01 & 1.13e+00 & 2.20e+04 & 1.20e+04 & 1.77e+02 & 2.94e+01 \\
    Quadratic & 4.75e+00 & 1.71e+00 & 1.87e-01 & 1.84e-01 & 6.54e-01 & 5.72e-01 & \best 2.06e+03 & \best 1.10e+03 & 1.06e+02 & 2.47e+01 \\
    RandomForest & 4.42e+00 & 5.62e-01 & \statsimilar 8.03e-02 & \statsimilar 6.30e-02 & 1.51e-01 & 1.59e-01 & 8.27e+03 & 4.96e+03 & \best 7.21e+01 & \best 1.75e+01 \\
    RBF & 3.67e+00 & 1.05e+00 & 1.33e-01 & 6.83e-02 & 7.21e-01 & 5.60e-01 & 2.87e+03 & 1.18e+03 & 1.18e+02 & 2.12e+01 \\
\bottomrule
  \end{tabular}
  }
  \label{tbl:synthetic_results:usb}
  \end{table}

\begin{table}[H]
\centering
\caption{Mean function performance using the EI (\textit{left table}) and UCB
         (\textit{right table}) acquisition functions. Median regret 
		 (\textit{column left}) and median absolute deviation from the median 
		 (MAD, \textit{column right}) after 200 function evaluations across the 51
		 runs are shown in paired columns. The method with the lowest median 
		 performance is shown in dark grey, with those statistically equivalent in
		 light grey.}
\begin{minipage}{0.49\textwidth}%
  \setlength{\tabcolsep}{2pt}
  \resizebox{\textwidth}{!}{%
  \begin{tabular}{l Sv Sv}
    \toprule
    \bfseries Mean function
    & \multicolumn{2}{c}{\bfseries \textsc{push4} (4)} 
    & \multicolumn{2}{c}{\bfseries \textsc{push8} (8)} \\ 
    & \multicolumn{1}{c}{Median} & \multicolumn{1}{c}{MAD}
    & \multicolumn{1}{c}{Median} & \multicolumn{1}{c}{MAD}  \\ \midrule
    Arithmetic & 1.53e-01 & 1.16e-01 & \statsimilar 2.77e+00 & \statsimilar 1.93e+00 \\
    Median & 1.54e-01 & 6.53e-02 & \best 2.19e+00 & \best 1.58e+00 \\
    Min & 2.39e-01 & 1.60e-01 & 3.53e+00 & 1.44e+00 \\
    Max & 1.28e-01 & 9.75e-02 & \statsimilar 2.66e+00 & \statsimilar 1.83e+00 \\
    Linear & 2.62e-01 & 2.10e-01 & 3.39e+00 & 2.05e+00 \\
    Quadratic & 1.92e-01 & 1.28e-01 & \statsimilar 3.16e+00 & \statsimilar 1.83e+00 \\
    RandomForest & \best 6.56e-02 & \best 5.71e-02 & \statsimilar 3.07e+00 & \statsimilar 2.24e+00 \\
    RBF & 1.64e-01 & 1.19e-01 & 3.35e+00 & 2.29e+00 \\
\bottomrule
  \end{tabular}
  }%
\end{minipage}\hfill%
\begin{minipage}{0.49\textwidth}%
  \setlength{\tabcolsep}{2pt}
  \resizebox{\textwidth}{!}{%
  \begin{tabular}{l Sv Sv}
    \toprule
    \bfseries Mean function
    & \multicolumn{2}{c}{\bfseries \textsc{push4} (4)} 
    & \multicolumn{2}{c}{\bfseries \textsc{push8} (8)} \\ 
    & \multicolumn{1}{c}{Median} & \multicolumn{1}{c}{MAD}
    & \multicolumn{1}{c}{Median} & \multicolumn{1}{c}{MAD}  \\ \midrule
    Arithmetic & 3.42e-01 & 2.18e-01 & 3.95e+00 & 1.57e+00 \\
    Median & 5.01e-01 & 3.06e-01 & 3.64e+00 & 2.05e+00 \\
    Min & 4.36e-01 & 3.69e-01 & 4.37e+00 & 1.72e+00 \\
    Max & 4.25e-01 & 3.20e-01 & 4.57e+00 & 1.74e+00 \\
    Linear & 7.48e-01 & 5.47e-01 & 4.00e+00 & 2.70e+00 \\
    Quadratic & 4.69e-01 & 4.37e-01 & 3.79e+00 & 1.94e+00 \\
    RandomForest & \best 1.05e-01 & \best 6.92e-02 & \best 2.36e+00 & \best 2.11e+00 \\
    RBF & 1.39e-01 & 9.29e-02 & 3.66e+00 & 2.43e+00 \\
\bottomrule
  \end{tabular}
  }%
\end{minipage}
\end{table}

\begin{table}[H]
\caption{Mean function performance using the EI (\textit{left table}) and UCB
         (\textit{right table}) acquisition functions. Median regret 
		 (\textit{column left}) and median absolute deviation from the median 
		 (MAD, \textit{column right}) after 200 function evaluations across the 51
		 runs are shown in paired columns. The method with the lowest median 
		 performance is shown in dark grey, with those statistically equivalent in
		 light grey.}
\begin{minipage}{0.49\textwidth}%
  \centering
  \setlength{\tabcolsep}{2pt}
  \resizebox{.65\textwidth}{!}{%
  \begin{tabular}{l Sv Sv}
    \toprule
    \bfseries Mean function
    & \multicolumn{2}{c}{\bfseries PitzDaily (10)} \\ 
    & \multicolumn{1}{c}{Median} & \multicolumn{1}{c}{MAD}  \\ \midrule
    Arithmetic & \statsimilar 8.39e-02 & \statsimilar 1.12e-03 \\
    Median & 8.60e-02 & 3.23e-03 \\
    Min & \statsimilar 8.43e-02 & \statsimilar 1.39e-03 \\
    Max & \best 8.38e-02 & \best 1.38e-03 \\
    Linear & 8.50e-02 & 2.24e-03 \\
    Quadratic & 8.57e-02 & 3.07e-03 \\
    RandomForest & 8.84e-02 & 6.38e-03 \\
    RBF & 8.79e-02 & 6.23e-03 \\
\bottomrule
  \end{tabular}
  }%
\end{minipage}\hfill%
\begin{minipage}{0.49\textwidth}%
  \centering
  \setlength{\tabcolsep}{2pt}
  \resizebox{.65\textwidth}{!}{%
  \begin{tabular}{l Sv Sv}
    \toprule
    \bfseries Mean function
    & \multicolumn{2}{c}{\bfseries PitzDaily (10)} \\ 
    & \multicolumn{1}{c}{Median} & \multicolumn{1}{c}{MAD}  \\ \midrule
    Arithmetic & \statsimilar 8.41e-02 & \statsimilar 3.92e-04 \\
    Median & \statsimilar 8.40e-02 & \statsimilar 1.96e-04 \\
    Min & \best 8.39e-02 & \best 2.01e-04 \\
    Max & \statsimilar 8.40e-02 & \statsimilar 3.07e-04 \\
    Linear & \statsimilar 8.41e-02 & \statsimilar 3.98e-04 \\
    Quadratic & 8.43e-02 & 9.23e-04 \\
    RandomForest & 8.77e-02 & 4.69e-03 \\
    RBF & 8.96e-02 & 8.23e-03 \\
\bottomrule
  \end{tabular}
  }%
\end{minipage}
\end{table}

\section{GP Modelling Error Using Different Mean Functions}
In this section we present a short investigation into the modelling capability
of the Gaussian process (GP) when combined with different mean functions.
In order to quantify the modelling capability of a GP using a prior mean 
function on a test function $f : \Real^d \mapsto \Real$, we created GP models
of $f$ using training data taken from the optimisation runs, evaluated a
set of $N = 1000$ test locations, and calculated the model's normalised root
mean squared error (NRMSE). 

More precisely, for a given test problem $f$, a GP model using the mean 
function to be evaluated was created using the first $100$ locations 
(including initial LHS samples) for each of the mean function's $51$ 
optimisation runs using the EI acquisition function. A set of $N = 1000$
test locations $X = \{\bx_n\}_{n=1}^N$ were chosen via Latin hypercube sampling
and the model's prediction NRMSE between the true function values $f_n$ and
predicted values from the surrogate model $\hat{f_n}$ were calculated for each
of the $51$ models. The NRMSE is defined as 
\begin{equation}
\text{NRMSE} = \frac{\sqrt{ \frac{1}{N} \sum_{n=1}^N ( f_n - \hat{f_n} )^2 }} 
                    {f_{max} - f_{min}},
\end{equation}
where $f_{max}$ and $f_{min}$ are the largest and smallest values of the test
locations evaluated using the real function. The test locations $X$ were paired
across models corresponding to optimisation runs using the same training data
so that a statistical comparison could be undertaken.

\begin{table}[t]
\centering
\resizebox{0.8\textwidth}{!}{%
\begin{tabular}[t]{l cccccccc}
               & Arith & Med & Min & Max & Lin & Quad & RF & RBF \\
\toprule
Branin         & 0.271 & 0.271 & 0.271 & 0.263 & 0.270 & 0.263 & \statsimilar 0.121 & \best 0.114 \\
Eggholder      & 0.217 & \statsimilar 0.165 & 0.216 & 0.177 & \best 0.161 & \statsimilar 0.163 & 0.198 & \statsimilar 0.166 \\
GoldsteinPrice & 0.136 & 0.136 & 0.136 & 0.136 & 0.136 & 0.136 & \best 0.069 & 0.136 \\
SixHumpCamel   & 0.232 & 0.231 & 0.231 & 0.223 & 0.215 & 0.216 & \best 0.142 & 0.228 \\
Shekel         & 0.265 & 0.215 & 0.494 & \statsimilar 0.058 & 0.083 & 0.094 & 0.142 & \best 0.053 \\
Ackley         & 0.663 & 0.716 & 0.316 & 0.899 & \best 0.107 & 0.378 & 0.334 & \statsimilar 0.117 \\
Hartmann6      & 0.222 & 0.394 & 0.184 & 0.154 & 0.140 & 0.185 & 0.195 & \best 0.097 \\
Michalewicz    & 0.271 & 0.311 & 0.417 & 0.316 & 0.205 & \statsimilar 0.193 & 0.360 & \best 0.190 \\
Rosenbrock     & 0.362 & 0.368 & 0.371 & 0.270 & 0.257 & \best 0.151 & 0.355 & \statsimilar 0.157 \\
StyblinskiTang & 0.325 & 0.347 & 0.360 & \best 0.151 & 0.267 & \statsimilar 0.199 & 0.220 & \statsimilar 0.168 \\
\pushfour      & 0.202 & 0.200 & 0.256 & 0.254 & \statsimilar 0.153 & \statsimilar 0.152 & 0.222 & \best 0.151 \\
\pusheight     & 0.205 & 0.202 & 0.238 & 0.168 & 0.136 & \statsimilar 0.129 & 0.191 & \best 0.125
\end{tabular}
}%
\caption{Median NRMSE modelling prediction error taken over across $51$
         GP models using the the $8$ evaluated mean functions on the ten
         synthetic and two robot pushing functions. The models with the lowest
         NRMSE are shown in dark grey, with those statistically equivalent to
         them shown in light grey.}
\label{tbl:rmse}
\end{table}
Table~\ref{tbl:rmse} shows the median NRMSE over the $51$ GP models created
using each mean function for the synthetic and robot pushing test problems.
The method with the lowest median NRMSE on each function is highlighted in 
dark grey, and those highlighted in light grey are statistically equivalent
to the best method according to a one-sided paired Wilcoxon signed-rank test
with Holm-Bonferroni correction ($p\geq0.05$). The RBF mean function provides
the best (or equivalent to the best) modelling error for the evaluated test 
locations on 10 of the 12 evaluated functions. Interestingly, this does not
correspond the same performance in Bayesian optimisation (recall Table~1 in the
main paper). We suspect this is because the experiments are evaluating
modelling of the entire function, whereas BO only requires the GP model to
reflect the true function within the region that the minima reside. 

Further work in quantifying the modelling error in BO could be carried out by
only calculating the NRMSE within a small hypercube surrounding the global 
optimum. However, the size of the hypercube required to accurately calculate the
modelling error of the important features for BO may vary between problems and
therefore additional evaluations may be required of sets of test data sampled
from hypercubes of different sizes.